\pdfoutput=1

\documentclass[11pt]{article}

\usepackage[final]{acl}

\usepackage{times}
\usepackage{latexsym}

\usepackage[T1]{fontenc}

\usepackage[utf8]{inputenc}

\usepackage{microtype}

\usepackage{inconsolata}

\usepackage{graphicx}
\usepackage{caption}
\usepackage{subcaption}

\usepackage{multirow}

\usepackage{amsmath}
\usepackage{amssymb}

\usepackage{xspace}
\usepackage{booktabs}
\usepackage{array}
\usepackage{geometry}
\usepackage{ulem}
\usepackage{color}

\definecolor{blue}{RGB}{0 ,138, 212}
\definecolor{blue_back}{RGB}{198, 229, 245}
\definecolor{green}{RGB}{0, 161, 134}
\definecolor{green_back}{RGB}{204, 236, 231}
\definecolor{gray}{RGB}{230, 230, 230}
\definecolor{red}{RGB}{246, 117, 106}
\definecolor{red_back}{RGB}{253, 227, 225}
\definecolor{black}{RGB}{0, 0, 0}
\newcommand{\saul}{\textsc{SAUL}\xspace}
\newcommand{\saulFT}{\textsc{SAUL}\xspace}

\newcommand{\blue}[1]{\textcolor{black}{#1}}

\title{Better Call \saul: Fluent and Consistent Language Model Editing \\ with Generation Regularization}

\author{
Mingyang Wang$^{1,2,3}$ \hspace*{0.2cm}
{Lukas Lange$^{1}$}\hspace*{0.2cm} 
{Heike Adel$^{4}$} \\
 {\bf Jannik Str\"{o}tgen$^{5}$ \hspace*{0.2cm} Hinrich Sch\"{u}tze$^{2,3}$} \\
  $^1$Bosch Center for Artificial Intelligence (BCAI) \hspace*{0.2cm}
  $^2$LMU Munich \\
  $^3$Munich Center for Machine Learning (MCML) \\
  $^4$Hochschule der Medien, Stuttgart\\
  $^5$Karlsruhe University of Applied Sciences \\
  \texttt{mingyang.wang2@de.bosch.com} 
  }

\def\mathindent{\mbox{\hspace{1.2cm}}}

\begin{document}
\maketitle
\begin{abstract}

To ensure large language models contain up-to-date knowledge, they need to be updated regularly. However, model editing is challenging as it might also affect knowledge that is unrelated to the new data. 
State-of-the-art methods identify parameters associated with specific knowledge and then modify them via direct weight updates. However, these locate-and-edit methods suffer from heavy computational overhead and lack theoretical validation. In contrast, directly fine-tuning the model on requested edits 
affects the model's behavior on unrelated knowledge, and significantly damages the model's generation fluency and consistency.
To address these challenges, we propose \saul, a streamlined model editing method that uses \textbf{s}entence concatenation with \textbf{au}gmented random facts for generation regu\textbf{l}arization. 
Evaluations on three model editing benchmarks show that \saul is a practical and reliable solution for model editing outperforming state-of-the-art methods while maintaining generation quality and reducing computational overhead. 
\end{abstract}

\section{Introduction}


Large Language Model (LLMs) have been shown to implicitly store factual knowledge in their parameters \cite{petroni-etal-2019-language, roberts-etal-2020-much}. However, since our world is changing, facts can become obsolete or incorrect. Thus, there is the need for \textit{model editing}, i.e., updating or fixing incorrect knowledge stored in LLMs without disrupting their overall functionality, in particular, leaving unrelated knowledge unchanged and keeping their generation quality on a high level.

\begin{figure}[t]
    \centering
    \includegraphics[width=1\linewidth]{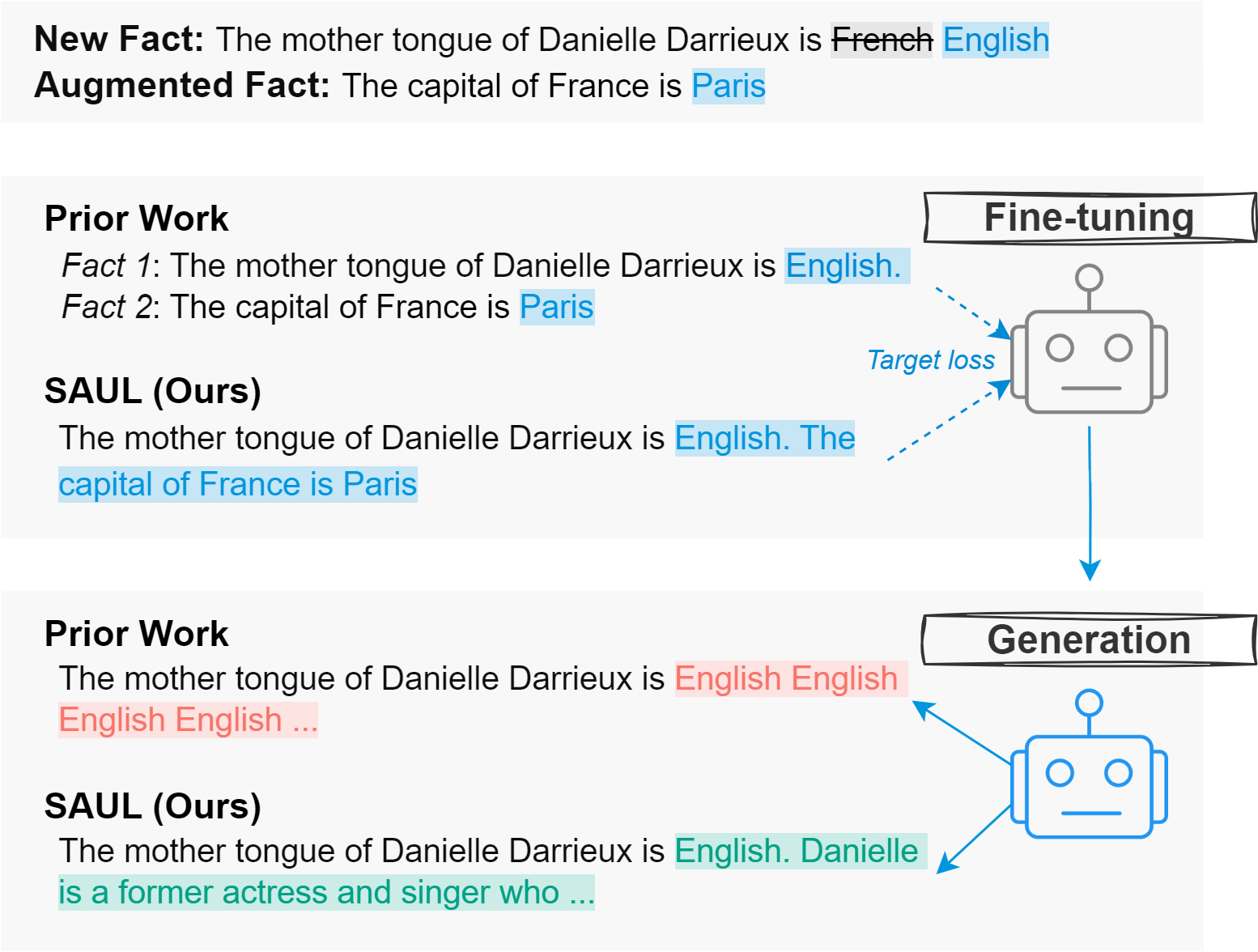}
    \caption{Comparison between \saul\ and prior work for model editing. 
    Prior work causes \color{red}{generation repetition}\color{black}{, as the fine-tuning loss focuses only on a few} \color{blue}{target tokens}\color{black}{. In contrast, \saul\ regularizes the model's generation with sentence concatenation. Consequently, the model can still generate }\color{green}{fluent text}\color{black}{ after model editing.}
    }
    \label{fig:teaser-image}
\end{figure}

The state-of-the-art model editing strategy is \textit{locate-and-edit} \cite{meng2022locating, meng2022mass}. It first identifies the location of knowledge inside the LLMs, and then directly modifies the weights it identified. While effective in practice, it requires significant computational overhead \cite{meng2022locating, meng2022mass}, and relies on an the locality hypothesis of factual knowledge \cite{hase2024does}. In contrast, fine-tuning on requested edits is straightforward and agnostic to model architectures. However, naive fine-tuning has been shown to adversely affect the model's behavior on unrelated facts and impair the fluency and consistency of the model's generation \cite{meng2022mass, yao-etal-2023-editing, gangadhar2024model}.


To overcome these challenges, we propose \saul, a novel fine-tuning approach that uses \textbf{s}entence concatenation with \textbf{au}gmented random facts for generation regu\textbf{l}arization. Augmenting random facts effectively preserves the model’s knowledge of unrelated facts. In addition, concatenating the target factual sentence with a random factual sentence prevents the overfitting on the target token(s). This effectively avoids the generation of disfluent sentences 
-- as shown in Figure~\ref{fig:teaser-image}.

We evaluate our approach on three model editing benchmarks. The results demonstrate that \saul\ not only outperforms existing state-of-the-art methods in terms of model editing performance but also effectively preserves the fluency and consistency of the model's outputs. This makes our method both simple and efficient, providing a viable solution for practical and reliable model editing in LLMs.

\section{Related Work}
Model editing is a targeted approach to updating the knowledge stored in LLMs. Existing works can be categorized as follows: (1) \blue{\textit{Fine-tuning} is a simple and straightforward way to update model's knowledge. However, it often affects model's behavior on unrelated knowledge and can degrade the model's generation quality.} (2) \textit{Memory-based} methods introduce an external memory unit for requested edits \blue{and employ a retriever to extract the most relevant facts for model editing} \cite{mitchell2022memory, huang2023transformer, zheng-etal-2023-edit}. (3) \textit{Meta-learning} (``learning to learn") methods use a hypernetwork to learn the necessary model updates \blue{in response to specific data or tasks, enabling the model to quickly adapt to new data without retraining from scratch.} \cite{de-cao-etal-2021-editing, mitchell2021fast}. (4) \textit{Locate-and-edit} methods identify parameters associated with specific knowledge and modify them through direct parameter updates \cite{meng2022locating, meng2022mass}. 

Recent work \cite{gangadhar2024model} proposes a straightforward \textit{fine-tuning-based} model editing method with data augmentation, showing competitive performance, but leading to unexpected generation failures. \blue{In contrast, we propose generation regularization, combined with data augmentation, which achieves state-of-the-art model editing performance while preserving the model’s generation quality. Our method ensures that the edited model retains its ability to generate coherent and fluent text, making it broadly applicable in real-world applications.} 

\section{Method}
We propose \saul, a novel model editing method that regularizes the model's generation via sentence concatenation with augmented random facts.

\paragraph{Model Editing Problem Definition.}
LLMs have been shown to memorize factual knowledge \cite{petroni-etal-2019-language, roberts-etal-2020-much, kassner-etal-2021-multilingual}. We consider a fact to be a sentence $x_i$ that describes a subject-relation-object triple $(s_i, r_i, o_i)$ in natural language. 
A model $f_{\theta}$ should recall the object $o_i$ given given a natural language prompt $pr_i = pr(s_i, r_i)$ consisting the subject $s_i$ and relation $r_i$. We focus on mass-editing, i.e., editing a set of multiple facts at once. Given the set of requested edits $\mathcal{E} = \{(s_i, r_i, o_i)\}_{i=1}^N$, model editing aims to alter the model's behavior for facts within the editing scope $\mathcal{X}_e$, which encompasses $\mathcal{E}$ along with its equivalence neighborhood $N(\mathcal{E})$, while leaving its knowledge for out-of-scope examples, i.e. $ (s_i, r_i, o_i) \notin \mathcal{X}_e $, unchanged. 


\begin{table*}[htbp]
  \centering
  \footnotesize
  \scalebox{0.95}{
    \begin{tabular}{lcccccccc}
    \toprule
    \multicolumn{1}{c}{\multirow{2}[2]{*}{\textbf{Editor}}} & \multirow{2}[2]{*}{\textbf{Time} (/edit)} & \multicolumn{3}{c}{\textbf{CounterFact}} & \multicolumn{2}{c}{\textbf{ZsRE}} & \multicolumn{2}{c}{\textbf{WikiRecent}} \\
    \cmidrule(lr){3-5} \cmidrule(lr){6-7} \cmidrule(lr){8-9}
          &       & Score & Fluency & Consistency & Score & Fluency & Score & Fluency \\
    \midrule
    Original GPT-J$^{\dagger}$ & 0.0s   & 22.4  & 622.4 & 29.4  & 26.4  & 599.0 & 37.4  & 600.8 \\
    \midrule
    MEND$^{\dagger}$  & 0.003s & 23.1  & 618.4 & 31.1  & 20.0  & -    & -    & - \\
    ROME$^{\dagger}$  & 1.3s  & 50.3  & 589.6 & 3.3   & 2.6   & -    & 35.0  & - \\
    MEMIT$^{\dagger}$ & 0.7s   & 85.8  & \textbf{619.9} & \textbf{40.1}  & 50.7  & -    & 67.3  & - \\
    \midrule
    FT$^{\dagger}$ & 0.2s & 62.4  & 452.1 & 4.3   & 58.8  & 559.9 & 67.2  & 570.0 \\
    FT + R + P$^{\dagger}$ & 0.9s   & 86.5  & 352.0 & 5.2   & 62.0  & -    & 68.5  & - \\
    FT + R + P* & 1.1s   & 86.6  & 208.7 & 4.7   & \textbf{64.2}  & 591.5 & \textbf{70.1}  & 501.3 \\
    \saulFT & 0.4s   & \textbf{87.7}  & 600.7 & 31.0  & 63.6  & \textbf{620.7} & 69.7  & \textbf{560.6} \\
    \bottomrule
    \end{tabular}%
    }
  \caption{Summary of the model editing results on three benchmark datasets.
  We present the editing score, generation fluency and consistency, and the required time per edit for each method.
  \saul\ demonstrates strong performance in all these metrics across datasets, providing a robust and efficient solution for model editing. $^{\dagger}$ and * denote results taken from prior works and reproduced by us, respectively.\protect\footnotemark[5]}
  \label{tab:overall-res}%
\end{table*}%

\begin{table}[htbp]
  \centering
  \footnotesize
    \begin{tabular}{p{1.8cm}p{5cm}}
    \toprule
    \textbf{New Fact} & Inner Circle railway line can be found in \colorbox{gray}{\sout{Melbourne}} \colorbox{blue_back}{\color{blue}{Singapore}}. \\
    \midrule
    \textbf{Editor} & \textbf{Generation} \\
    \midrule
    Original GPT-J & Inner Circle railway line's surroundings include \color{green}{the following suburbs and areas…} \\
    FT + P + R & Inner Circle railway line's surroundings include \color{red}{\colorbox{red_back}{Melbourne} \colorbox{red_back}{Melbourne} \colorbox{red_back}{Melbourne} ...} \\
    \saulFT\ (Ours) & Inner Circle railway line's surroundings include \color{green}{residential areas. Inner Circle railway line can be found in \colorbox{green_back}{Singapore}... } \\
    \bottomrule
    \end{tabular}%
  \caption{Comparison of the model's generation after model editing. While FT+P+R fails to edit the knowledge and generates \color{red}{repetitive tokens}\color{black}{, \saul\ successfully incorporates the new fact into its }\color{green}{fluent generation}\color{black}{.}}
  \label{tab:res-generation}%
\end{table}%

\paragraph{Naive Fine-tuning for Model Editing.}
For a set of edits $\mathcal{E}$, fine-tuning-based methods
optimize
the conditional likelihood of the target object
given subject $s_i$ and relation $r_i$ of the
fact formulated as a natural language prompt $pr_i$: 
\\[0.25cm]
\mathindent$\underset{\theta}{\min} 
\underset{(s_i, r_i, o_i) \in \mathcal{E}}{\sum}
-\log p_{\theta}(o_i | s_i,
r_i)$
\\[0.05cm]


\paragraph{Random Fact Augmentation.} While naive fine-tuning has shown good editing efficacy, it harms generality and locality by not generalizing the edits to paraphrased sentences and altering the model’s predictions on unrelated facts \cite{meng2022mass}. \newcite{gangadhar2024model} demonstrate that fine-tuning with augmented paraphrases and random facts significantly improves generality and locality performance. Inspired by this work, we adopt the idea of data augmentation with random facts. We use random true facts from the training split provided by \newcite{gangadhar2024model}.\footnote{We do not use paraphrase fact augmentation as preliminary experiments showed a degradation of the model’s generation quality, which we will analyze in detail in Section~\ref{sec:res-da-ablation}.}

\paragraph{Generation Regularization.} We find that the post-edit model after fine-tuning leads to undesired generation failures, with the model generating repeating target tokens, as illustrated in Figure~\ref{fig:teaser-image}. 
We hypothesize that this occurs because the conditional likelihood-based optimization makes the model focus excessively on the target token(s), thus losing its general generation capability.
We propose to concatenate the factual sentence $x_i \in \mathcal{X}_e$ and the random factual sentence $a_j \in \mathcal{A}$ for fine-tuning.\footnote{\blue{In addition to concatenating random factual sentences to the factual sentence $x_i$, we explore other suffix options, including paraphrased sentences and combinations of both paraphrased and random sentences. See Table~\ref{tab:da-ablation} in Section~\ref{sec:res} for a more detailed discussion.}} Formally, \saul\ optimizes:
\\[0.25cm]
\mbox{\hspace{0.5cm}} $\underset{\theta}{\min} \underset{(s_i, r_i, o_i, a_j) \in \mathcal{E \cup A}}{\sum} -\log p_{\theta}(o_i, a_j \mid s_i, r_i) $
\\[0.25cm]
The sentence concatenation strategy regularizes the model’s generation, so that it maintains the model’s generation quality and still produces fluent natural sentences after editing.

\begin{table*}[t]
  \centering
  \footnotesize
  \scalebox{0.9}{
    \begin{tabular}{lccccccc}
    \toprule
    \multicolumn{1}{c}{\multirow{2}[2]{*}{\textbf{Editor}}} & \multicolumn{3}{c}{\textbf{CounterFact}} & \multicolumn{2}{c}{\textbf{ZsRE}} & \multicolumn{2}{c}{\textbf{WikiRecent}} \\
    \cmidrule(lr){2-4} \cmidrule(lr){5-6} \cmidrule(lr){7-8}
          & Score & Fluency & Consistency & Score & Fluency & Score & Fluency \\
    \midrule
    Original GPT-J & 22.4  & 622.4 & 29.4  & 26.4  & 599.0 & 37.4  & 600.8 \\
    \midrule
    FT 21st & 57.0  & \textbf{584.4} & \textbf{14.9}  & 37.9  & \textbf{566.4} & 45.7  & \textbf{595.8} \\
    FT 3-8th & 60.8  & 553.8 & 8.7   & 56.7  & 549.5 & \textbf{69.2}  & 574.3 \\
    FT all & \textbf{62.4}  & 452.1 & 4.3   & \textbf{58.8}  & 559.9 & 67.2  & 570.0 \\
    FT LoRA & 55.4  & 494.4 & 5.7   & 57.8  & 543.9 & 67.5  & 546.8 \\
    \midrule
    \saulFT 3-8th & \textbf{89.8}  & 595.4 & 30.1  & \textbf{63.6}  & 615.0 & 69.4  & \textbf{587.9} \\
    \saulFT all & 87.7  & \textbf{600.7} & \textbf{31.0}  & \textbf{63.6}  & \textbf{620.7} & \textbf{69.7}  & 560.6 \\
    \bottomrule
    \end{tabular}%
    }
  \caption{We compare fine-tuning on different layers of the language model. 
  Applying \saul\ on different layers achieves notable improvements, demonstrating its effectiveness across various fine-tuning paradigms.}
  \label{tab:res-ft-compare}%
\end{table*}%

\begin{table*}[htbp]
    \centering
    \begin{minipage}[b]{0.7\textwidth}
    \centering
    \footnotesize
    \scalebox{0.9}{
        \begin{tabular}{lccccccc}
        \toprule
        \multicolumn{1}{c}{\multirow{2}[2]{*}{\textbf{Editor}}} & \multicolumn{3}{c}{\textbf{CounterFact}} & \multicolumn{2}{c}{\textbf{ZsRE}} & \multicolumn{2}{c}{\textbf{WikiRecent}} \\
        \cmidrule(lr){2-4} \cmidrule(lr){5-6} \cmidrule(lr){7-8}
              & Score & Fluency & Consistency & Score & Fluency & Score & Fluency \\
        \midrule
        Original GPT-J & 22.4  & 622.4 & 29.4  & 26.4  & 599.0 & 37.4  & 600.8 \\
        \midrule
        FT    & 62.4  & 452.1 & 4.3   & 58.8  & 559.9 & 67.2  & 570.0 \\
        FT + R & 85.3  & 379.0 & 3.5   & 58.6  & 564.2 & 69.8  & 454.6 \\
        FT + P & 70.7  & 190.9 & 5.6   & 63.7  & 607.2 & 69.0  & 541.5 \\
        FT + P + R & 86.6  & 208.7 & 4.7   & \textbf{64.2}  & 591.5 & 70.1  & 501.3 \\
        \midrule
        \saulFT w/ R & \textbf{87.7}  & \textbf{600.7} & \textbf{31.0}  & 63.6  & \textbf{620.7} & 69.7  & \textbf{560.6} \\
        \saulFT w/ P & 68.7  & 366.8 & 8.6   & 54.4  & 466.9 & 69.5  & 406.4 \\
        \saulFT w/ P + R & 87.5  & 447.6 & 18.0  & 63.5  & 490.3 & \textbf{70.5}  & 437.8 \\
        \bottomrule
        \end{tabular}%
        }
        \caption{We investigate different data augmentation strategies. Our method, \saul\ with random augmentation, shows the best overall performance across datasets in terms of editing scores, generation fluency and consistency. }
        \label{tab:da-ablation}
    \end{minipage}\hfill
    \begin{minipage}[b]{0.28\textwidth}
        \centering
        \includegraphics[width=\linewidth]{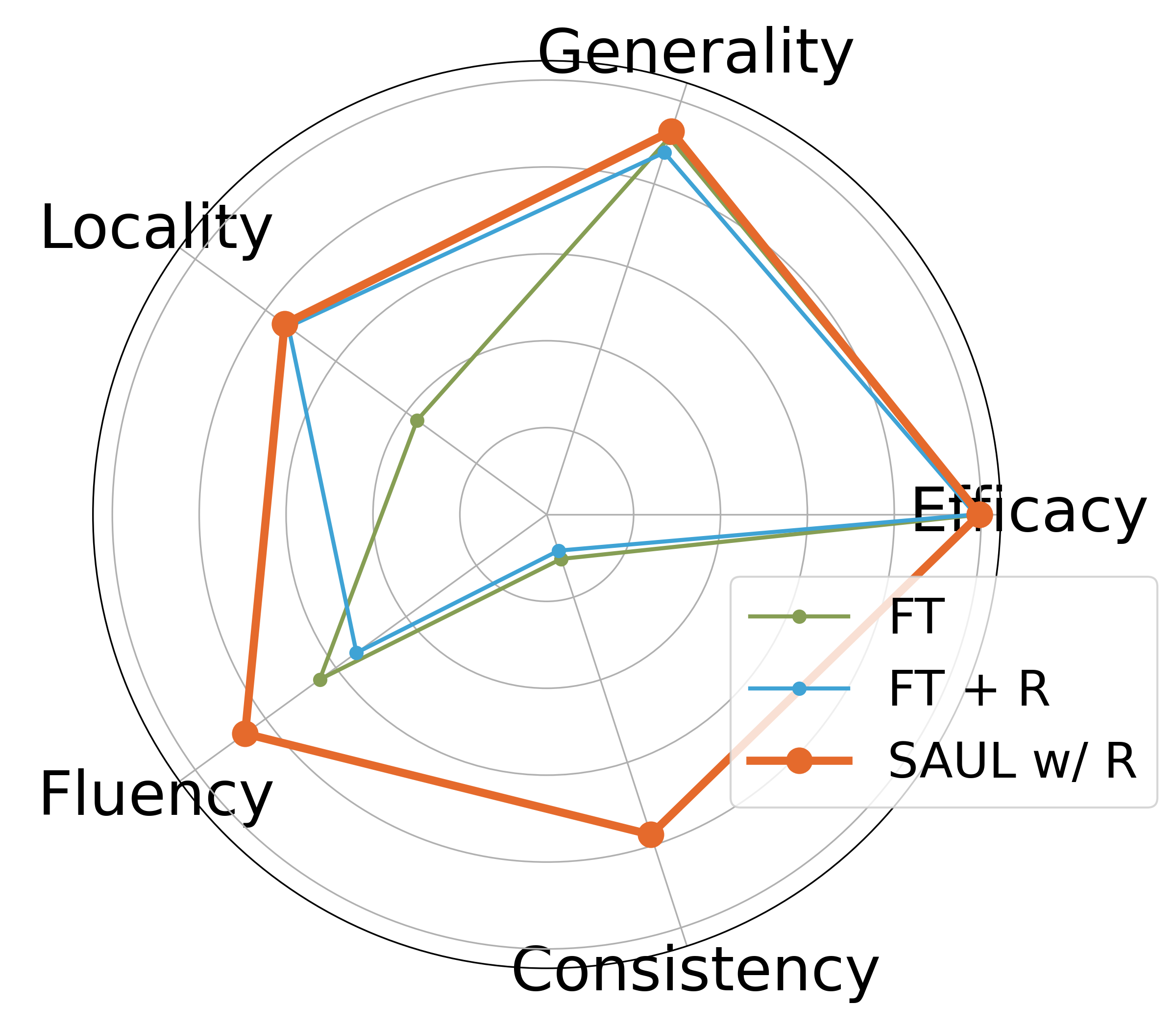}
        \captionof{figure}{Comparison of naive fine-tuning, fine-tuning with random augmentation, and \saul.}
        \label{fig:da-ablation}
    \end{minipage}
\end{table*}

\section{Experimental Setup}
\label{sec:exp-setup}

\paragraph{Datasets and Baselines.} We evaluate \saul and related methods on three datasets: CounterFact \cite{meng2022locating}, ZsRE \cite{levy-etal-2017-zero}, and WikiRecent \cite{cohen2024evaluating}.\footnote[3]{Details of these dataset are provided in Appendix~\ref{sec:appendix-dataset}} 

We include the following baselines: MEND \cite{mitchell2021fast} - a hypernetwork-based method; ROME \cite{meng2022locating} and MEMIT \cite{meng2022mass} - locate-and-edit methods; FT and FT+R+P
\cite{gangadhar2024model} - fine-tuning without and with data augmentation, respectively.\footnote[4]{R: random augmentation, P: paraphrase augmentation.} \blue{Please refer to Appendix~\ref{sec:appendix-additional-res} for a comparison between these gradient-based methods and the parameter-free in-context knowledge editing method (IKE) \cite{zheng-etal-2023-edit}.}

\footnotetext[5]{FT+R+P* in Section~\ref{sec:res} refers to the reproduction results we obtained by fine-tuning all model layers; Prior work (FT+R+P) use LoRA \cite{hu2022lora} for fine-tuning}

\paragraph{Training Details.} 
We follow the mass-editing setting as in \newcite{meng2022mass, gangadhar2024model}.
For each edit, we augment $N_r$ unrelated true facts provided by \newcite{gangadhar2024model} for sentence concatenation. We fine-tune all model layers of GPT-J 6B \cite{gpt-j} and compare different fine-tuning paradigms in Section~\ref{sec:res-ft-ablation}. 

\paragraph{Evaluation Metrics.} Model editing performance is evaluated by three metrics: (1) \textit{Efficacy} measures if the model predicts the new target $o_i$ with a greater probability than the original prediction $o_i^\text{-}$. 
(2) \textit{Generality} evaluates if the post-edit model can generalize to an equivalent paraphrase of the edit sentence. 
(3) \textit{Locality} assesses the accuracy on the knowledge out of the edit scope $\mathcal{X}_e$. 

Besides, 
we report \textit{fluency} and \textit{consistency} 
following prior work \cite{meng2022locating, meng2022mass, gangadhar2024model}. For fluency, we calculate the n-gram entropy of the model's generated text.\footnote[6]{We provide examples and analysis of the generation fluency in Section~\ref{sec:res-overall}.}
For consistency, we compare the generated text
with reference texts about subjects sharing the target property. The consistency score is the cosine similarity between their unigram TF-IDF vectors.\footnote[7]{We only report the consistency score on the CounterFact Dataset as this is the only dataset with reference texts.}

We calculate the harmonic mean of efficacy, generality, and locality as the \textit{editing score} following prior works. We report this editing score, along with fluency and consistency in Section~\ref{sec:res}. We provide the complete results in Appendix~\ref{sec:appendix-additional-res}.

\section{Results and Analysis}
\label{sec:res}

\paragraph{Overall Results.}
\label{sec:res-overall}
As shown in Table~\ref{tab:overall-res}, \saul\ consistently demonstrates strong performance in terms of editing score, generation quality, and computational efficiency. 
In particular, it performs better than the state-of-the-art, but complex MEMIT system on all evaluation datasets.
While FT+R+P achieves competitive editing scores, it shows poor generation quality, suggesting that the model's generation quality has been damaged during editing. 

In Table~\ref{tab:res-generation}, we provide a qualitative comparison of the model's generation after editing. We observe that FT+R+P fails to incorporate the new fact and overfits to the target token, leading to repetitive generation of ``Melbourne''. However, \saul\ maintains the generation quality and successfully integrates the new fact into the generated text.

\paragraph{Ablation Study: Fine-tuning Paradigms.}
\label{sec:res-ft-ablation}
We compare naive fine-tuning (no augmentation) and \saul\ on different layers of GPT-J and using LoRA for parameter-efficient fine-tuning. Our selection of fine-tuning layers is based on conclusions from previous locate-and-edit works: \newcite{meng2022locating} find that fine-tuning the 21st layer of GPT-J yields the best performance, while \newcite{meng2022mass} identify layers 3 to 8 as the most critical layers for factual recall. 

The experimental results in Table~\ref{tab:res-ft-compare} show that fine-tuning on layers 3-8 and all layers achieves strong editing scores. 
While \saulFT 3-8th shows the highest score on CounterFact, \saulFT all performs best on the other two datasets. We suspect this is because \newcite{meng2022mass} use CounterFact for parameter localization, and layers 3-8 might not generalize well to other datasets. In contrast, our method is dataset-agnostic and consistently improves performance across various datasets.

\paragraph{Ablation Study: Data Augmentation.}
\label{sec:res-da-ablation}
We study different data augmentation strategies for model editing.\footnote[8]{We follow the data augmentation strategies used in \newcite{gangadhar2024model}.} We experiment with naïve fine-tuning, i.e., no augmentation, along with fine-tuning and \saul\ with random augmentation (R), paraphrase augmentation (P), and both augmentations (P+R). 

As shown in Table~\ref{tab:da-ablation}, fine-tuning with any data augmentation significantly improves the editing score compared to naive fine-tuning, but at the cost of generation quality. In particular, paraphrase augmentation causes a degradation of the model's generation quality, likely because it introduces unnatural sentence segments.\footnote[9]{Please refer to Appendix~\ref{sec:appendix-dataset} for more details.} As shown in Figure~\ref{fig:da-ablation}, our method, \saulFT w/ R, outperforms other methods in terms of generation fluency and consistency, and achieving strong editing scores across datasets.

\section{Conclusion}

In this work, we proposed \saul, a novel fine-tuning method to address the challenges of preserving unrelated knowledge in LLMs and maintaining high generation quality during model editing. To achieve this, \saul\ regularizes the generation process through sentence concatenation with augmented random facts. Our evaluation on three benchmark datasets demonstrated that \saul\ outperforms state-of-the-art methods while maintaining generation quality and reducing computational overhead. Consequently, \saul\ offers an efficient and practical solution for model editing in LLMs.

\section*{Limitations}
\paragraph{Data Augmentation Strategies. } 
Data augmentation is an active research area in natural language processing. In this work, we explore paraphrase and random augmentation to regularize the model’s generation. Investigating additional data augmentation strategies could further improve performance and offer new insights into the model editing task, which we leave for future work.

\paragraph{Multilingual Model Editing Evaluation.} 
Our evaluations are limited to monolingual datasets due to the absence of well-established multilingual datasets. To assess the effectiveness and generalizability of \saul\ across diverse linguistic contexts, experiments with multilingual datasets are essential. This would help determine how well our method adapts to languages with various vocabulary sets and linguistic features.

\paragraph{Experiments with Different Numbers of Edits.}
In this work, we focus on the mass-editing setting following prior works \cite{meng2022mass, gangadhar2024model}. Specifically, the CounterFact, ZsRE, and WikiRecent datasets used in this work provide 10,000, 10,000, and 1,266 requested edits, respectively. Investigating the performance and stability of \saul\ under varying numbers of edits could provide valuable information about its scalability. This would be an interesting direction for future research.

\section*{Ethical Considerations}

One potential ethical issue of this work arises from the use of the CounterFact dataset which contains incorrect factual knowledge. While this dataset is valuable for testing and improving model editing methods, it inherently introduces the risk of propagating incorrect information if not carefully managed.
Model editing based on such a dataset can inadvertently lead to the generation of incorrect information and hallucinated text. 

\section*{Acknowledgements}
We would like to thank the anonymous reviewers for their constructive feedback. This work was partially supported by Deutsche Forschungsgemeinschaft (project SCHU 2246/14-1).

\bibliography{anthology,custom}

\clearpage
\appendix

\section{ Appendix}
\label{sec:appendix}

\subsection{Dataset Information}
\label{sec:appendix-dataset}
We evaluate \saul\ and related methods on three datasets: CounterFact, ZsRE, and WikiRecent.\footnote[10]{We select the datasets following previous works \cite{mitchell2021fast, meng2022locating, meng2022mass, gangadhar2024model}, and leave the extension to other model editing datasets, such as \newcite{zhong-etal-2023-mquake, ammar-khodja-etal-2024-wikifactdiff, nie2024bmike}, for future work.} CounterFact \cite{meng2022mass} is a dataset that includes artificially created counterfacts to test the ability of model editing methods to add counterfactual information to the language model. ZsRE \cite{levy-etal-2017-zero} is a question-answering dataset consisting of 10,000 real-world facts, used to test model editing methods for adding correct information.  WikiRecent \cite{cohen2024evaluating} collects factual knowledge that has been inserted into WikiData after July 2022. 

Specifically, the CounterFact, ZsRE, and WikiRecent datasets provide 10,000, 10,000, and 1,266 requested edits, respectively. For each requested edit, we augment 20 unrelated true facts provided by \newcite{gangadhar2024model} for sentence concatenation. For the data augmentation ablation study, we add paraphrase samples for augmentation following \newcite{gangadhar2024model}. They augment the paraphrase data by generating free texts using the GPT-J model and prepend these texts to the original factual sentence for model editing. The generated sentence segments are listed in Table~\ref{tab:paraphrase-exp}. As discussed in Section~\ref{sec:res-da-ablation}, paraphrase augmentation causes a degradation in the model's generation quality, likely because it introduces unnatural sentences such as ''Q: How can I use a. The mother tongue of Danielle Darrieux is English".

\begin{table}[htbp]
  \centering
  \footnotesize
  \scalebox{0.9}{
    \begin{tabular}{l}
    \toprule
    \textbf{Paraphrase prefix} \\
    \midrule
    ''Q: . "\\
    ''Q: . "\\
    ''The present invention relates."  \\
    ''The role of the."  \\
    ''\textbackslash{}n \textbackslash{}n-."  \\
    ''Q: Why is my code not."  \\
    ''Q: What is the correct way."  \\
    ''The present invention relates in general to the manufacture."  \\
    ''The role of the family in the development of."  \\
    ''\textbackslash{}n \textbackslash{}n-\textbackslash{}n \textbackslash{}n1\textbackslash{}n."  \\
    ''A new report from the Center for Immigration Studies."  \\
    ''Q: How can I use a."  \\
    ''Q: How to use multiple variables."  \\
    ''\textbackslash{}n \textbackslash{}n=\textbackslash{}n \textbackslash{}n1\textbackslash{}n."  \\
    ''Q: What is the difference in."  \\
    \bottomrule
    \end{tabular}%
    }
  \caption{Examples of the prefix text used for paraphrase augmentation.}
  \label{tab:paraphrase-exp}%
\end{table}%

\subsection{Implementation Details}
\label{sec:appendix-implementation}
We use the AdamW optimizer \cite{adamw} for all experiments.Table \ref{tab:hyperparams} provides detailed hyperparameter choices for \saul\ across datasets. The training
was performed on Nvidia A100 GPUs.\footnote[11]{All experiments ran on a carbon-neutral GPU cluster.}

\begin{table}[htbp]
\footnotesize
  \centering
  \scalebox{0.95}{
    \begin{tabular}{lccc}
    \toprule
          & CounterFact & ZsRE  & WikiRecent \\
    \midrule
    Epochs & \multicolumn{3}{c}{40} \\
    Early stop patience & \multicolumn{3}{c}{5} \\
    Batch size & \multicolumn{3}{c}{32} \\
    No. augmented facts & 20    & 20    & 10 \\
    Learning rate & 5e-5 & 2e-5 & 1e-4 \\
    \bottomrule
    \end{tabular}%
    }
  \caption{Hyperparameters used on three model editing datasets used in this work.}
  \label{tab:hyperparams}%
\end{table}%


\subsection{Additional Experimental Results}
\label{sec:appendix-additional-res}
As introduced in Section~\ref{sec:exp-setup}, model editing performance is evaluated using \textit{efficacy}, \textit{generality}, and \textit{locality}. In Section~\ref{sec:res}, we report the harmonic mean of these three metrics in the main paper for brevity. Here in Table~\ref{tab:complete-counterfact} to \ref{tab:complete-da-ablation-wikirecent}, we provide the complete evaluation results, including all these model editing metrics and the generation metrics \textit{fluency} and \textit{consistency}.
\blue{Here, we also include the experimental results of the in-context knowledge editing (IKE) method \cite{zheng-etal-2023-edit}, which allows the model to acquire new knowledge directly from the input context \cite{NEURIPS2020_1457c0d6, zhang-etal-2024-impact}. \footnote[12]{We experiments use the IKE implementation in EasyEdit \cite{wang-etal-2024-easyedit}.} It is important to note that our work focuses on gradient-based model editing in the mass-editing setting, where multiple facts are edited simultaneously. In contrast, knowledge editing with in-context learning is limited to the single-edit case, making it an unsuitable baseline for our approach. Nonetheless, we include the IKE results to offer a more comprehensive comparison and to highlight \saul's relative strengths.}

\begin{table*}[htbp]
  \centering
  \footnotesize
    \begin{tabular}{lcccccc}
    \toprule
    \multicolumn{1}{c}{\multirow{2}[4]{*}{\textbf{Editor}}} & \multicolumn{6}{c}{\textbf{CounterFact}} \\
\cmidrule{2-7}          & Score & \textit{Efficacy} & \textit{Generalit}y & \textit{Locality} & Fluency & Consistency \\
    \midrule
    Original GPT-J & 22.4  & 15.2  & 17.7  & 83.5  & 622.4 & 29.4 \\
    \midrule
    MEND  & 23.1  & 15.7  & 18.5  & 83.0 & 618.4 & 31.1 \\
    ROME  & 50.3  & 50.2  & 50.4  & 50.2  & 589.6 & 3.3 \\
    MEMIT & 85.8  & 98.9  & 88.6  & 73.7  & \textbf{619.9} & \textbf{40.1} \\
    \blue{IKE} & 74.3  & 100.0  & 95.1  & 50.3  & 620.9 & 29.2 \\
    FT + R + P & 86.5  & 98.8  & 93.6  & 72.0  & 352.0 & 5.2 \\
    FT + R + P* & 86.6  & 98.1  & 95.1 & 71.8  & 208.7 & 4.7 \\
    \saulFT  & \textbf{87.7} & 99.6 & 92.8  & 74.8  & 600.7 & 31.0 \\
    \bottomrule
    \end{tabular}%
  \caption{Complete evaluation results on CounterFact of \saul\ and related methods on three benchmark datasets.}
  \label{tab:complete-counterfact}%
\end{table*}%
\begin{table*}[htbp]
  \centering
  \footnotesize
  \scalebox{1}{
    \begin{tabular}{lccccc}
    \toprule
    \multicolumn{1}{c}{\multirow{2}[4]{*}{\textbf{Editor}}} & \multicolumn{5}{c}{\textbf{ZsRE}} \\
\cmidrule{2-6}          & Score & \textit{Efficacy} & \textit{Generality} & \textit{Locality} & Fluency \\
    \midrule
    Original GPT-J & 26.4  & 26.4  & 25.8  & 27.0  & 599.0 \\
    \midrule
    MEND  & 20.0  & 19.4  & 18.6  & 22.4  & - \\
    ROME  & 2.6   & 21.0  & 19.6  & 0.9   & - \\
    MEMIT & 50.7  & 96.7  & 89.7  & 26.6  & - \\
    \blue{IKE} & \textbf{65.4}  & 100.0  & 98.7  & 38.9  & 584.6 \\
    FT + R + P & 62.0  & 99.9. & 97.0 & 35.6  & - \\
    FT + R + P* & 64.2 & 97.0  & 87.2  & 40.1 & 591.5 \\
    \saulFT & 63.6  & 99.9 & 93.4  & 37.8  & \textbf{620.7} \\
    \bottomrule
    \end{tabular}%
    }
  \caption{Complete evaluation results on ZsRE of \saul\ and related methods on three benchmark datasets.}
  \label{tab:complete-zsre}%
\end{table*}%

\begin{table*}[htbp]
  \centering
  \footnotesize
    \begin{tabular}{lccccc}
    \toprule
    \multicolumn{1}{c}{\multirow{2}[4]{*}{\textbf{Editor}}} & \multicolumn{5}{c}{\textbf{WikiRecent}} \\
\cmidrule{2-6}          & Score & \textit{Efficacy} & \textit{Generality} & \textit{Locality} & Fluency \\
    \midrule
    Original GPT-J & 37.4  & 34.4  & 34.5  & 45.3  & 600.8 \\
    \midrule
    MEND  & -    & -    & -    & -    & - \\
    ROME  & 35.0  & 39.8  & 25.5  & 46.9 & - \\
    MEMIT & 67.3  & 99.2  & 80.2  & 45.3  & - \\
    \blue{IKE}  & \textbf{77.8}  & 100.0  & 85.4  & 54.3 & \textbf{574.5} \\
    FT + R + P & 68.5  & 99.6 & 84.6  & 45.8  & - \\
    FT + R + P* & 70.1 & 99.6 & 93.4 & 45.4  & 501.3 \\
    \saulFT & 69.7  & 99.5  & 89.1  & 46.0  & 560.6 \\
    \bottomrule
    \end{tabular}%
  \caption{Complete evaluation results on WikiRecent of \saul\ and related methods on three benchmark datasets.}
  \label{tab:complete-wikirecent}%
\end{table*}%

\begin{table*}[htbp]
  \centering
  \footnotesize
    \begin{tabular}{lcccrcc}
    \toprule
    \multicolumn{1}{c}{\multirow{2}[4]{*}{\textbf{Editor}}} & \multicolumn{6}{c}{\textbf{CounterFact}} \\
\cmidrule{2-7}          & Score & \textit{Efficacy} & \textit{Generality} & \multicolumn{1}{c}{\textit{Locality}} & Fluency & Consistency \\
    \midrule
    Original GPT-J & 22.4  & 15.2  & 17.7  & 83.5  & 622.4 & 29.4 \\
    \midrule
    FT 21st & 57.0  & 84.3  & 52.0  & 46.5 & \textbf{584.4} & \textbf{14.9} \\
    FT 3-8th & 60.8  & 99.9  & 82.5  & 36.8  & 553.8 & 8.7 \\
    FT all & \textbf{62.4} & 99.9  & 91.2 & 36.9  & 452.1 & 4.3 \\
    FT LoRA & 55.4  & 100.0 & 71.6  & 33.1  & 494.4 & 5.7 \\
    \midrule
    \saulFT\ 3-8th & \textbf{89.8} & 99.5  & 92.4  & 79.7 & 595.4 & 30.1 \\
    \saulFT\ all & 87.7  & 99.6 & 92.8 & 74.6  & \textbf{600.7} & \textbf{31.0} \\
    \bottomrule
    \end{tabular}%
  \caption{Complete evaluation results on CounterFact for the ablation study with various fine-tuning paradigms.}
  \label{tab:complete-ft-ablation-counterfact}%
\end{table*}%

\begin{table*}[htbp]
  \centering
  \footnotesize
  \caption{Complete evaluation results on ZsRE for the ablation study with various fine-tuning paradigms.}
    \begin{tabular}{lccccc}
    \toprule
    \multicolumn{1}{c}{\multirow{2}[4]{*}{\textbf{Editor}}} & \multicolumn{5}{c}{\textbf{ZsRE}} \\
\cmidrule{2-6}          & Score & \textit{Efficacy} & \textit{Generality} & \textit{Locality} & Fluency \\
    \midrule
    Original GPT-J & 26.4  & 26.4  & 25.8  & 27.0  & 599.0 \\
    \midrule
    FT 21st & 37.9  & 45.7  & 43.4  & 29.2  & \textbf{566.4} \\
    FT 3-8th & 56.7  & 98.9  & 96.5 & 30.9  & 549.5 \\
    FT all & \textbf{58.8} & 99.5 & 96.3  & 32.7 & 559.9 \\
    FT LoRA & 57.8  & 96.5  & 92.4  & 32.6  & 543.9 \\
    \midrule
    \saulFT\ 3-8th & \textbf{63.6} & 99.7  & 85.1  & 39.4 & 615.0 \\
    \saulFT\ all & \textbf{63.6} & 99.9 & 93.4 & 37.8  & \textbf{620.7} \\
    \bottomrule
    \end{tabular}%
  \caption{Complete evaluation results on ZsRE for the ablation study with various fine-tuning paradigms.}
  \label{tab:complete-ft-ablation-zsre}%
\end{table*}%

\begin{table*}[htbp]
  \centering
  \footnotesize
    \begin{tabular}{lccccc}
    \toprule
    \multicolumn{1}{c}{\multirow{2}[4]{*}{\textbf{Editor}}} & \multicolumn{5}{c}{\textbf{WikiRecent}} \\
\cmidrule{2-6}          & Score & \textit{Efficacy} & \textit{Generality} & \textit{Locality} & Fluency \\
    \midrule
    Original GPT-J & 37.4  & 34.4  & 34.5  & 45.3  & 600.8 \\
    \midrule
    FT 21st & 45.7  & 48.8  & 43.7  & 45.0  & \textbf{595.8} \\
    FT 3-8th & \textbf{69.2} & 99.6 & 87.8 & 45.5 & 574.3 \\
    FT all & 67.2  & 99.6 & 79.8  & 45.3  & 570.0 \\
    FT LoRA & 67.5  & 99.4  & 81.4  & 45.3  & 546.8 \\
    \midrule
    \saulFT\ 3-8th 3-8th & 69.4  & 99.5 & 85.5  & 46.5 & \textbf{587.9} \\
    \saulFT\ 3-8th all & \textbf{69.7} & 99.5 & 89.1 & 46.0  & 560.6 \\
    \bottomrule
    \end{tabular}%
  \caption{Complete evaluation results on WikiRecent for the ablation study with various fine-tuning paradigms.}
  \label{tab:complete-ft-ablation-wikirecent}%
\end{table*}%

\begin{table*}[htbp]
  \centering
  \footnotesize
    \begin{tabular}{lcccccc}
    \toprule
    \multicolumn{1}{c}{\multirow{2}[4]{*}{\textbf{Editor}}} & \multicolumn{6}{c}{\textbf{CounterFact}} \\
\cmidrule{2-7}          & Score & \textit{Efficacy} & \textit{Generality} & \textit{Locality} & Fluency & Consistency \\
    \midrule
    Original GPT-J & 22.4  & 15.2  & 17.7  & 83.5  & 622.4 & 29.4 \\
    \midrule
    FT    & 62.4  & 99.9  & 91.2  & 36.9  & 452.1 & 4.3 \\
    FT + R & 85.3  & 98.7  & 87.6  & 73.5  & 379.0 & 3.5 \\
    FT + P & 70.7  & 99.9  & 99.2  & 44.7  & 190.9 & 5.6 \\
    FT + P + R & 86.6  & 98.1  & 95.1  & 71.8  & 208.7 & 4.7 \\
    \saul\ w/ R & \textbf{87.7}  & 99.6  & 92.8  & 74.6  & \textbf{600.7} & \textbf{31.0} \\
    \saul\ w/ P & 68.7  & 100.0 & 97.4  & 42.7  & 366.8 & 8.6 \\
    \saul\ w/ P + R & 87.5  & 99.8  & 92.1  & 74.5  & 447.6 & 18.0 \\
    \bottomrule
    \end{tabular}%
  \caption{Complete evaluation results on CounterFact for the ablation study with various data augmentation strategies.}
  \label{tab:complete-da-ablation-counterfact}%
\end{table*}%

\begin{table*}[htbp]
  \centering
  \footnotesize
    \begin{tabular}{lccccc}
    \toprule
    \multicolumn{1}{c}{\multirow{2}[4]{*}{\textbf{Editor}}} & \multicolumn{5}{c}{\textbf{ZsRE}} \\
\cmidrule{2-6}          & Score & \textit{Efficacy} & \textit{Generality} & \textit{Locality} & Fluency \\
    \midrule
    Original GPT-J & 26.4  & 26.4  & 25.8  & 27.0  & 599.0 \\
    \midrule
    FT    & 58.8  & 99.5  & 96.3  & 32.7  & 559.9 \\
    FT + R & 58.6  & 99.6  & 98.5  & 32.2  & 564.2 \\
    FT + P & 63.7  & 99.8  & 94.2  & 37.8  & 607.2 \\
    FT + P + R & \textbf{64.2}  & 97.0  & 87.2  & 40.1  & 591.5 \\
    \saul\ w/ R & 63.6  & 99.9  & 93.4  & 37.8  & \textbf{620.7} \\
    \saul\ w/ P & 54.4  & 99.9  & 96.0  & 28.8  & 466.9 \\
    \saul\ w/ P + R & 63.5  & 99.9  & 94.9  & 37.4  & 490.3 \\
    \bottomrule
    \end{tabular}%
  \caption{Complete evaluation results on ZsRE for the ablation study with various data augmentation strategies.}
  \label{tab:complete-da-ablation-zsre}%
\end{table*}%

\begin{table*}[ht]
  \centering
  \footnotesize
    \begin{tabular}{lccccc}
    \toprule
    \multicolumn{1}{c}{\multirow{2}[4]{*}{\textbf{Editor}}} & \multicolumn{5}{c}{\textbf{WikiRecent}} \\
\cmidrule{2-6}          & Score & \textit{Efficacy} & \textit{Generality} & \textit{Locality} & Fluency \\
    \midrule
    Original GPT-J & 37.4  & 34.4  & 34.5  & 45.3  & 600.8 \\
    \midrule
    FT    & 67.2  & 99.6  & 79.8  & 45.3  & \textbf{570.0} \\
    FT + R & 69.9  & 99.6  & 92.2  & 45.4  & 454.6 \\
    FT + P & 69.0  & 99.5  & 85.4  & 46.1  & 541.5 \\
    FT + P + R & 70.1  & 99.6  & 93.4  & 45.4  & 501.3 \\
    \saul\ w/ R & 69.7  & 99.5  & 89.1  & 46.0  & 560.6 \\
    \saul\ w/ P & 69.5  & 99.5  & 87.7  & 46.1  & 406.4 \\
    \saul\ w/ P + R & \textbf{70.5}  & 99.5  & 86.7  & 47.7  & 437.8 \\
    \bottomrule
    \end{tabular}%
  \caption{Complete evaluation results on WikiRecent for the ablation study with various data augmentation strategies.}
  \label{tab:complete-da-ablation-wikirecent}%
\end{table*}%

\end{document}